\documentclass{article}

\usepackage[preprint]{neurips_2024}

\usepackage[utf8]{inputenc} 
\usepackage[T1]{fontenc}    
\usepackage{hyperref}       
\usepackage{url}            
\usepackage{booktabs}       
\usepackage{amsfonts}       
\usepackage{nicefrac}       
\usepackage{microtype}      
\usepackage{xcolor}         

\usepackage{graphicx}
\usepackage{caption}
\usepackage{subcaption}
\usepackage{natbib}
\usepackage{comment}

\usepackage[draft,textsize=footnotesize,textwidth=15mm]{todonotes}

\title{The Evolution of Multimodal Model Architectures}

\author{
  Shakti N.~Wadekar \\
  Purdue University \\
  \texttt{swadekar@purdue.edu} \\
  \And
  Abhishek~Chaurasia \\
  Chaos Industries Inc. \\
  \texttt{abhi@choasinc.com} \\
   \And
  Aman~Chadha \\
  Stanford; Amazon\thanks{Work does not relate to position at Amazon.} \\
  \texttt{hi@aman.ai} \\
  \And
  Eugenio~Culurciello \\
  Purdue University\\
  \texttt{euge@purdue.edu} \\
}

\begin{document}

\maketitle

\begin{figure}[ht]
  \centering
  \includegraphics[width=1.0\textwidth]{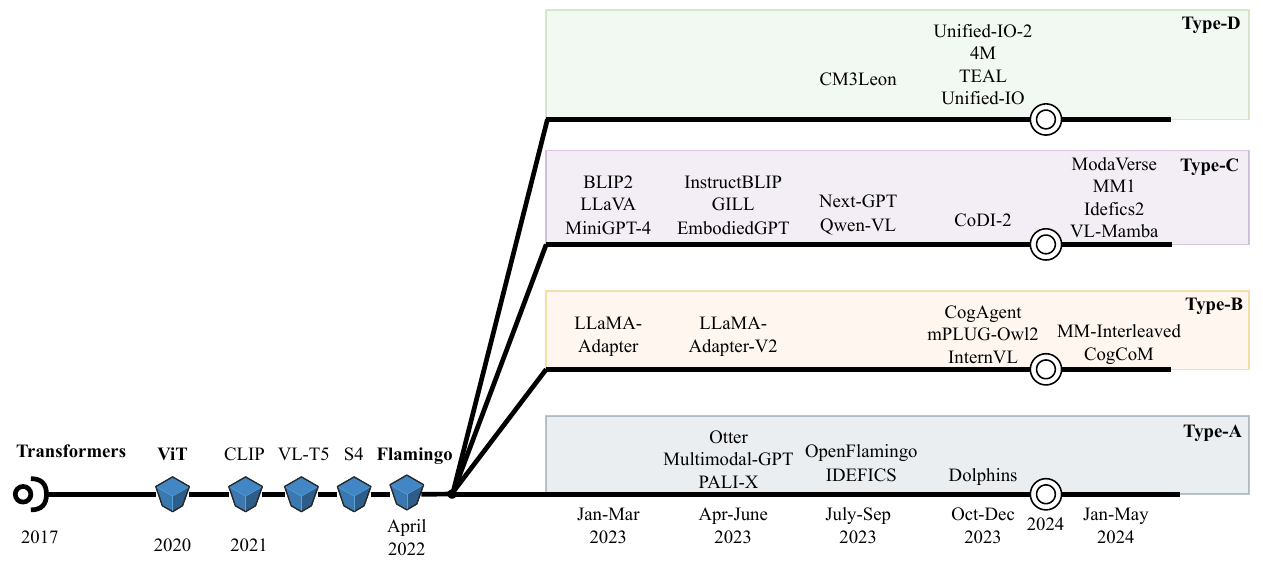}
  \caption{Development timeline of Multimodal models grouped in four proposed architecture types.}
  \label{fig:developement_timeline}
\end{figure}

\begin{abstract}
  This work uniquely identifies and characterizes four prevalent multimodal model architectural patterns in the contemporary multimodal landscape.
  Systematically categorizing models by architecture type facilitates monitoring of developments in the multimodal domain.
  Distinct from recent survey papers that present general information on multimodal architectures, this research conducts a comprehensive exploration of architectural details and identifies four specific architectural types.
  The types are distinguished by their respective methodologies for integrating multimodal inputs into the deep neural network model.
  The first two types (Type A and B) deeply fuses multimodal inputs within the internal layers of the model, whereas the following two types (Type C and D) facilitate early fusion at the input stage.
  Type-A employs standard cross-attention, whereas Type-B utilizes custom-designed layers for modality fusion within the internal layers.
  On the other hand, Type-C utilizes modality-specific encoders, while Type-D leverages tokenizers to process the modalities at the model’s input stage.
  The identified architecture types aid the monitoring of any-to-any multimodal model development.
  Notably, Type-C and Type-D are currently favored in the construction of any-to-any multimodal models.
  Type-C, distinguished by its non-tokenizing multimodal model architecture, is emerging as a viable alternative to Type-D, which utilizes input-tokenizing techniques. 
  To assist in model selection, this work highlights the advantages and disadvantages of each architecture type based on data and compute requirements, architecture complexity, scalability, simplification of adding modalities, training objectives, and any-to-any multimodal generation capability.
\end{abstract}

\begin{figure}
  \centering
  \includegraphics[width=0.96\textwidth]{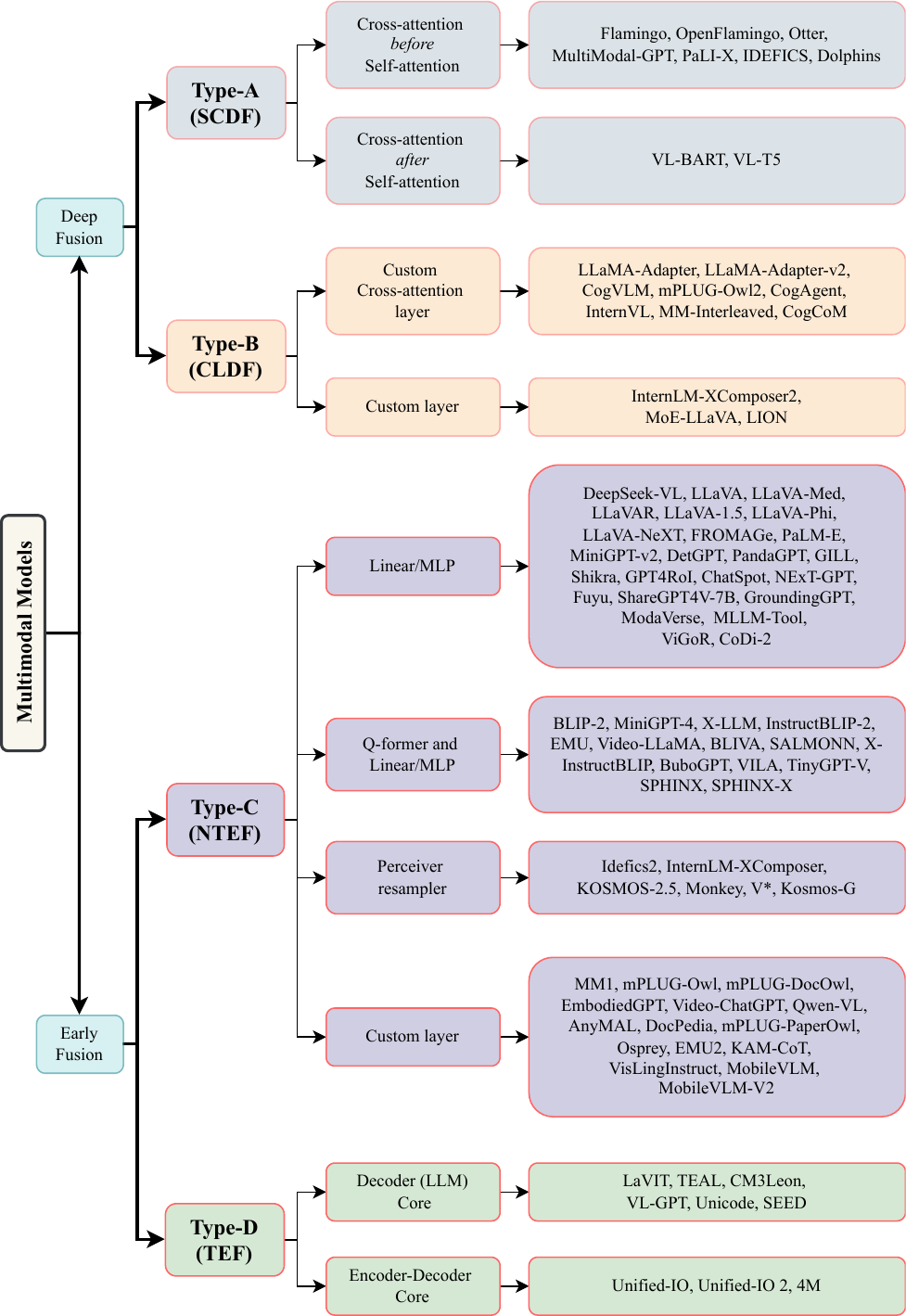}
  \caption{Taxonomy of multimodal model architectures. Four distinct types of multimodal architectures and their sub-types are outlined. Various models are systematically catalogued to the types and sub-types. Deep Fusion: Type-A and Type-B fuses multimodal inputs within the internal layers of the model. Early Fusion: Type-C and Type-D facilitate fusion at the input stage. Type-A uses standard cross-attention, whereas Type-B utilizes custom-designed cross-attention or specialized layers. Type-C is a non-tokenizing multimodal model architecture, while Type-D, employs input-tokenization (discrete tokens). SCDF: Standard Cross-attention based Deep Fusion. CLDF: Custom Layer based Deep Fusion. NTEF: Non-Tokenized Early Fusion. TEF: Tokenized Early Fusion.}
  \label{fig:types_subtypes}
\end{figure}

\section{Introduction}

The multimodal domain of machine learning has seen significant advancements in recent years. The proliferation of models capable of processing images, audio, or video in conjunction with text (language) has notably expanded
(\cite{alayrac2022flamingo}, \cite{lu2023unified}, \cite{mizrahi20244m}, \cite{wu2023next}, \cite{yang2024teal}, \cite{tang2023codi}).
Remarkable strides have been particularly evident in the integration of image and text modalities across diverse vision-language tasks, primarily because of the Transformer model \cite{vaswani2017attention}.
The Transformer model \cite{vaswani2017attention}, a pioneering deep neural network (NN) architecture, has spearheaded a unified framework for cross-domain learning. This singular model exhibits remarkable efficacy in comprehending and processing data from diverse domains.
The introduction of the Transformer model \cite{vaswani2017attention} for Natural Language Processing (NLP) in 2017 marked the inception of transformer-based model architectures.
Subsequently, the introduction of the Vision Transformer (ViT) \cite{dosovitskiy2021image} and CLIP \cite{radford2021learning} for the vision domain showcased the versatility of transformers in handling image-related tasks.
This demonstration highlighted the transformer's ability to learn from diverse domains, prompting a series of initiatives aimed at constructing models capable of jointly processing image and text data, leveraging the robust Transformer model architecture.
Flamingo \cite{alayrac2022flamingo}, a transformer-based multimodal model that incorporates both image and text data as input, exhibited outstanding performance on vision-language tasks. These results served as a catalyst for further advancements in the multimodal domain and encouraged the integration of additional modalities \cite{awadalla2023openflamingo}, \cite{li2023otter}, \cite{gong2023multimodal}, \cite{laurenccon2024obelics}, \cite{ma2023dolphins}.

Diverse methodologies have been employed in numerous research efforts dedicated to handling mixed modalities,
including augmenting Large Language Models (LLMs) to create multimodal architectures \cite{alayrac2022flamingo}, \cite{gong2023multimodal}, \cite{zhang2023llama}, \cite{gao2023llama}, \cite{wang2023cogvlm}, \cite{ye2023mplug}, \cite{chen2023internvl}, \cite{tian2024mm}, \cite{lin2024moe}, training encoder-decoder style transformers with different input modalities \cite{mizrahi20244m}, \cite{lu2022unified}, \cite{lu2023unified}, and exploring alternative approaches (Section \ref{sec:relatedwork}).
The plethora of research presents challenges in effectively monitoring the progression of model architectures and identifying emerging trends in next-generation multimodal model designs.
We examine contemporary landscape of state-of-the-art multimodal models, and identify distinct multimodal model architectures based on the fusion of inputs into the deep neural networks.
Primarily, we group existing multimodal architectures in four broad categories namely: Type - A, B, C, and D.
In Type-A and Type-B architectures, deep fusion of input modalities is realized through the integration of inputs within the internal layers of the model, whereas in Type-C and Type-D architectures, early fusion of modalities occurs at the input stage of the model.
Details of these four types are discussed in Section \ref{section:MultimodalAI}.
Summary of our contributions:
\begin{itemize}
    \item To the best of our knowledge, this is the first work that explicitly identifies the four broad architecture types: Type - A, B, C, and D. Figure \ref{fig:types_subtypes} shows the taxonomy of multimodal model architectures. We associate state-of-the-art models with these types, and outline their advantages and disadvantages, thereby facilitating a simplified comprehension, visualization and selection of multimodal model architectures.
    \item Furthermore, our work also underscores the principal architectural types involved in constructing any-to-any modality multimodal models, which cannot be found in other survey works like \cite{zhang2024mmllms}, \cite{yin2024survey}, \cite{caffagni2024r}, \cite{wang2023large}, \cite{wu2023multimodal}, \cite{wu2023multimodal}, \cite{guo2023survey}.
    \item To facilitate model selection, this study highlights the advantages and disadvantages of each architecture type, considering factors such as training data and compute requirements, architecture complexity, scalability, ease of integrating modalities, and any-to-any modality capability.
\end{itemize}

\section{Related Work}
\label{sec:relatedwork}

In this section, we list and discuss existing survey literature encompassing multimodal learning, multimodal data, and multimodal large language models (Table \ref{table:survey-work}).
Our work illuminates a spectrum of multimodal architecture types through an examination of numerous multimodal works, which is absent in other survey literature.

\begin{table}[ht]
  \caption{List of survey works related to recent multimodal developments.}
  \label{table:survey-work}
  \centering
  \begin{tabular}{ll}
    \toprule
    \textbf{Multimodal Survey Article}      & \textbf{Year} \\
    \toprule
    MM-LLMs: Recent Advances in MultiModal LLMs \cite{zhang2024mmllms}   & 2024     \\
    A Survey on Multimodal Large Language Models \cite{yin2024survey}   & 2024     \\
    The (R)Evolution of Multimodal LLMs: A Survey  \cite{caffagni2024r}    & 2024      \\
    Large-scale Multi-modal Pre-trained Models: A Survey \cite{wang2023large}          & 2023  \\
    Multimodal Large Language Models: A Survey \cite{wu2023multimodal}           & 2023  \\
    Multimodal Learning With Transformers: A Survey \cite{xu2023multimodal}            & 2023  \\
    A Survey on Image-text Multimodal Models  \cite{guo2023survey}          & 2023  \\
    
    \bottomrule
  \end{tabular}
\end{table}

\textbf{Multimodal LLMs:}
\cite{zhang2024mmllms}, examines recent advancements in Multimodal Large Language Models (MLLMs), and explores NNs created by enhancing LLMs with mixed modality.
It assesses the performance of mainstream MLLMs across 18 vision-language benchmarks. 
While it offers a comprehensive overview of the general architecture of MLLMs, it notably overlooks the critical inclusion of Type-D \ref{section:type4} multimodal model architecture.
Type-D \ref{section:type4} is an emerging and popular multimodal model architecture type for developing any-to-any modality models.
\cite{yin2024survey}, similar to \cite{zhang2024mmllms}, offers insights into the intricate details of typical multimodal model architecture. However, it too lacks discussion on the Type-D multimodal architecture.
It thoroughly presents details of pretraining, finetuning, and alignment methods, as well as the data used for Multimodal Large Language Models (MLLMs).
\cite{caffagni2024r}, shows a general multimodal model architecture, provides a comprehensive inventory of the components present in multimodal architectures, encompassing a diverse range of LLM variants, vision encoders, and vision-to-language connectors/adapters.
It also compares these SOTA multimodal models on 14 multimodal benchmarks.
Multimodal models tailored to specific domains, such as document understanding, medical vision learning, autonomous driving, and embodied AI, are discussed, but it too noticeably lacks information about Type-D multimodal architecture.

\textbf{Multimodal foundational models, training tasks, data and challenges:}
Foundational multimodal models are explored in \cite{wu2023multimodal}.
It extensively describes multimodal tasks and its related datasets for training the foundational multimodal models. 
\cite{wu2023multimodal} lacks details about the architectures of these models.
The survey work \cite{wang2023large}, contains details of the models till 2022 and reviews multimodal pretraining datasets, pretraining tasks, pretrained model architectures and downstream multimodal tasks.
Since the work only explores models till 2022, it lacks details of multimodal model architectures, which are currently prevalent.
\cite{baltruvsaitis2018multimodal}, highlights challenges in multimodal machine learning.
It enumerates five multimodal challenges: representation, translation, alignment, fusion, and co-learning.
Discussion about different types of model architectures is absent in this work.
\cite{xu2023multimodal}, analyzes transformer model architecture from the perspective of multimodal learning.
It lists various self-attention variants for input multimodal data/embeddings fusion.
Based on the pretraining loss function, it categorizes the pretraining tasks used for multimodal learning with transformers.
Similar to \cite{baltruvsaitis2018multimodal}, it highlights challenges of multimodality fusion and alignment.
\cite{xu2023multimodal} too fails to discuss about different types of multimodal model architectures.
\cite{guo2023survey}, specifically talks about models for image and text modalities.
It investigates developments that have directly or indirectly influenced the evolution of multimodal large language models.
It lists various image-text multimodal tasks and matches the widely used state-of-the-art models (till October 2023) with each of these tasks.
The multimodal tasks include image captioning, visual reasoning, visual grounding and text-to-image generation.

\section{Multimodal Model Architectures: A Taxonomy}
\label{section:MultimodalAI}

The fusion of multimodal inputs into the deep neural network through various methods results in a range of architectural configurations.
This study analyzes model architectures with mixed modalities and categorizes them into four distinct types based on the fusion of modalities.
Two overarching categories are discernible: \textit{Deep Fusion}, wherein the fusion of modalities occurs within the internal layers of the model, and \textit{Early Fusion}, characterized by the fusion of modalities at the model's input.
Within each category, we observe two primary clusters. 
In the domain of Deep Fusion, the integration of modalities with internal layers manifests in: Type-A which employs standard cross-attention layers, and Type-B which utilizes custom-designed layers. 
Conversely, in the domain of Early Fusion, multimodal inputs take two principal forms: non-tokenized\footnote{No discrete tokenization} multimodal inputs as Type C, and discretely tokenized multimodal inputs as Type-D. 
These inputs are directly supplied to the input of the transformer model for early fusion, which can be either a decoder-only or an encoder-decoder style.
Therefore, we define four distinct types of multimodal model architectures within the current landscape of multimodal models: Type-A \ref{section:type1}, Type-B \ref{section:type2}, Type-C \ref{section:type3} and Type-D \ref{section:type4}.

Section \ref{section:MultimodalAI} comprehensively outlines each architecture type, including information on their training data and computational requirements.
Figure \ref{fig:types_subtypes} maps together the multimodal model architecture types and the corresponding SOTA multimodal models.
Figure \ref{fig:developement_timeline} depicts the timeline of multimodal model development.
The 4 types, Type-A, Type-B, Type-C and Type-D are described in Section \ref{section:type1}, \ref{section:type2}, \ref{section:type3}, \ref{section:type4} respectively.
Advantages and disadvantages of each multimodal model architecture type is listed in Section \ref{section:advantages_all}.

\subsection{Type-A: Standard Cross-Attention based Deep Fusion (SCDF)}
\label{section:type1}

\begin{figure}[ht]
  \centering
  \includegraphics[width=0.9\textwidth]{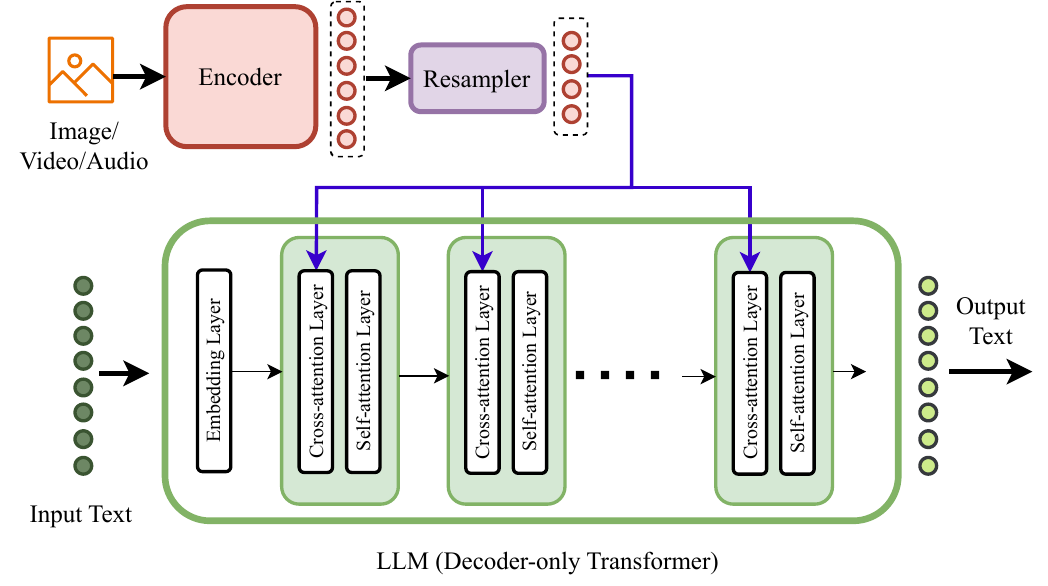}
  \caption{Type-A multimodal model architecture. The input modalities are deeply fused into the internal layers of the LLM using standard cross-attention layer. The cross-attention can be added either before (sub-type A.1) or after (sub-type A.2) the self-attention layer. Modality-specific encoders process the different input modalities. A resampler is used to output a fixed number of modality (visual/audio/video) tokens, given a variable number of input tokens at the input.}
  \label{fig:type1_architecture}
\end{figure}

Type-A architecture mostly comprise of early multimodal models.
Figure \ref{fig:type1_architecture} shows a general Type-A model architecture.
It typically involves a pre-trained LLM and integrating standard cross-attention layers into its internal architecture to achieve deep fusion of input modalities.
The multimodal data (image/audio/video) is fed through modality specific encoders.
A resampler is used to generate a fixed number of tokens that aligns with the requirements of the decoder layer.
These resampler outputs are then directed to the internal layers of the LLM using cross-attention layers.
The Type-A models exhibit a dichotomy in this aspect -- the cross-attention layer can be added before or after the self-attention layer in the model's architecture. 
This pre- and post-introduction of cross-attention w.r.t. self-attention in the LLM, results in the emergence of two distinct model architecture subtypes within Type-A.
Section \ref{section:type1subtype1} and \ref{section:type1subtype2} details more about each sub-type.
Models belonging to this architecture type include Flamingo \cite{alayrac2022flamingo}, OpenFlamingo (open-source replication of Flamingo) \cite{awadalla2023openflamingo}, Otter (trained on MIMIC-IT dataset on top of OpenFlamingo) \cite{li2023otter}, MultiModal-GPT (derived from OpenFlamingo) \cite{gong2023multimodal}, PaLI-X \cite{chen2023pali}, IDEFICS (open-access reproduction of Flamingo) \cite{laurenccon2024obelics}, Dolphins (based on OpenFlamingo architecture) \cite{ma2023dolphins}, VL-BART \cite{cho2021unifying} and VL-T5 \cite{cho2021unifying}.
At present, models with this architecture commonly engage in processing image and text modalities, subsequently producing textual outputs.
Pretraining necessitates a substantial volume of data samples and computational resources, as observed in the resource-intensive implementations of Flamingo, OpenFlamingo, PaLI-X, and IDEFICS.
Fine-tuning and/or instruction tuning these NNs can be achieved with minimal computational resources, a characteristic shared by the multimodal architectures discussed in this study.
This trend is evident in the Otter, Multimodal-GPT, and Dolphins models, where fine-tuning or instruction tuning is exclusively performed using specific data, leveraging limited computational resources (typically less than or equal to 8 A100 GPUs).
The comparative advantages and disadvantages of the Type-A multimodal model architecture, in relation to Types B, C, and D, are detailed in Section\ref{section:type1advantages}.

\subsubsection{Subtype A.1}
\label{section:type1subtype1}

Figure \ref{fig:type1_architecture} illustrates the model architecture sub-type, featuring the cross-attention layers \textit{before} each self-attention layer within the decoder (LLM).
Models belonging to this sub-group include Flamingo \cite{alayrac2022flamingo}, OpenFlamingo \cite{awadalla2023openflamingo}, Otter \cite{li2023otter}, MultiModal-GPT \cite{gong2023multimodal}, PaLI-X \cite{chen2023pali}, IDEFICS \cite{laurenccon2024obelics} and Dolphins \cite{ma2023dolphins}.
Flamingo and its derivative multimodal models generally belong to this architectural sub-type.

\textit{\textbf{Training and data:}}
\textbf{Flamingo} model is trained with next-text-token prediction objective, where the input consists of interleaved image and text pairs and the model outputs text.
For pretraining, it uses M3W \cite{alayrac2022flamingo}, ALIGN \cite{jia2021scaling},  LTIP (Long Text \& Image Pairs) \cite{alayrac2022flamingo} and  VTP (Video \& Text Pairs) \cite{alayrac2022flamingo}.
For finetuning, VQAV2 \cite{antol2015vqa}, COCO \cite{chen2015microsoft}, VATEX \cite{wang2019vatex}, VizWiz \cite{gurari2018vizwiz}, MSRVTTQA \cite{xu2017video}, VisDial \cite{das2017visual}, YouCook2 \cite{zhou2018towards}, and TextVQA \cite{singh2019towards} datasets were used.
\textbf{OpenFlamingo} models are also pretrained with the next-text-token prediction objective using 60M interleaved (MMC4 \cite{zhu2024multimodal}) examples and 120M LAION-2B \cite{schuhmann2022laion} examples \cite{awadalla2023openflamingo}. 
It uses similar finetuning datasets as Flamingo.
\textbf{Otter} created MIMIC-IT \cite{li2023otter} dataset, and trained OpenFlamingo model on it.
This work improves OpenFlamingo's instruction-following and in-context learning ability.
\textbf{MultiModal-GPT} finetunes OpenFlamingo by adding LoRA \cite{hu2021lora} weights to the LLM layers.
This model too is trained using next-text-token prediction objective.
Language training datasets include Dolly-15k and Alpaca-GPT4 \cite{peng2023instruction}.
Vision-language datasets encompass A-OKVQA \cite{schwenk2022okvqa}, COCO Caption \cite{karpathy2015deep}, OCR-VQA \cite{mishra2019ocr} and data from LLaVA \cite{li2024llava} \& Mini-GPT4 \cite{zhu2023minigpt}.
\textbf{PaLI-X} can process image, text and video inputs.
Image and video inputs are encoded with ViT encoder.
The text inputs and encoded image \& video outputs are then processed by an encoder-decoder style transformer model.
Most of the pretraining tasks (with the exception of the masked image token task) predict text-only output from the multimodal input \cite{chen2023pali}.
Pretraining datasets include WebLI (image-text pairs) \cite{chen2022pali}, CC3M \cite{sharma2018conceptual} and VTP (video data) datasets.
Finetuning datasets include COCO (Karpathy split) \cite{karpathy2015deep},  NoCaps \cite{agrawal2019nocaps}, VQAv2 \cite{goyal2017making},  OKVQA \cite{marino2019ok}, TallyQA \cite{acharya2019tallyqa}, VizWizCap \cite{gurari2020captioning}, TextVQA \cite{singh2019towards}, STVQA \cite{biten2019scene}, OCRVQA \cite{mishra2019ocr}, InfoVQA \cite{mathew2022infographicvqa}, DocVQA \cite{mathew2021docvqa}, ChartQA \cite{masry2022chartqa}, MSR-VTT \cite{xu2016msr}, Activity-Net \cite{krishna2017dense}, VATEX \cite{wang2019vatex}, SMIT \cite{monfort2021spoken}, NExT-QA \cite{xiao2021next}.
\textbf{IDEFICS} was trained on Wikipedia (\cite{heafield2011kenlm}; \cite{laurenccon2022bigscience}), Public Multimodal Dataset \cite{singh2022flava}, LAION \cite{webster2023duplication}, and on a new 115B token dataset called OBELICS \cite{laurenccon2024obelics}.
It follows Flamingo architecture style. 
LLaMA \cite{touvron2023llama} is used as LLM and OpenClip as the vision encoder.
\textbf{Dolphins} is based on OpenFlamingo. 
It is trained on driving data.
This work utilizes BDD-X \cite{kim2018textual} to establish instruction dataset, focusing on four key AV (autonomous vehicle) tasks like behavior comprehension, control signal forecasting, behavior analysis, and in-depth conversation \cite{ma2023dolphins}.

\textit{\textbf{Compute resources:}} 
For \textbf{Flamingo}, all training and evaluations were performed on TPUv4 instances. The
largest model containing 80 billion parameters was trained on 1536 chips for 15 days and sharded across 16 device \cite{alayrac2022flamingo}.
\textbf{OpenFlamingo} was trained using 64 A100 GPUs.
While \textbf{Otter} only utilizes 1 A100 GPU or 4 RTX-3090 GPU, to instruction tune on MIMIC-IT \cite{li2023otter} dataset. 
The Otter models undergo solely instruction tuning without any pretraining process.
\textbf{MultiModal-GPT} uses 8 A100 GPUs for finetuning training. 
The MultiModal-GPT model does not undergo pretraining.
For \textbf{PaLI-X}, no resource detail are provided.
\textbf{IDEFICS}'s 9B-parameter models is trained on OBELICS-only \cite{laurenccon2024obelics} and LAION-only \cite{schuhmann2022laion} datasets using 32 A100 (80GB) GPUs, and on OBELICS + LAION using 64 (80GB) A100s, for approximately 6 days. 
IDEFICS's large model is trained using 512 80GB A100 GPUs \cite{laurenccon2024obelics}. 
\textbf{Dolphins} is fully trained using 4 NVIDIA A100 GPUs.

\subsubsection{Subtype A.2}
\label{section:type1subtype2}

This sub-type consists of a standard encoder-decoder transformer architecture, featuring a cross-attention layer placed \textit{after} each self-attention layer within the decoder.
Generally, language models follow a decoder-only structure. 
However, BART and T5 adopt an encoder-decoder style transformer architecture.
Unlike sub-type A.1, sub-type A.2 does not utilize resampler in its architecture.
Research work \cite{cho2021unifying}, extended these language models to build mixed modality models named VL-BART and VL-T5.

\textit{\textbf{Training and data:}}
VL-T5 and VL-BART are designed to handle input comprising of images and text, and produces textual outputs.
Visual embeddings from image are obtained using a CNN model.
The image embeddings and text tokens are given as input to the encoder of the BART/T5.
The model outputs text based on the task mentioned in the text input.
A standard next-text-token prediction objective is used for training.
Pretraining data includes,  MS-COCO (\cite{lin2014microsoft}; \cite{chen2015microsoft}), Visual Genome \cite{krishna2017visual}, VQA-v2.0 \cite{goyal2017making}, GQA balanced version \cite{hudson2019gqa}, and Visual7W \cite{zhu2016visual7w}.
Finetuning data encompasses  VQA \cite{goyal2017making}, GQA,  NLVR$^2$ \cite{suhr2018corpus}, RefCOCOg \cite{mao2016generation}, COCO Caption Karpathy \cite{karpathy2015deep} and Multi30K \cite{elliott2016multi30k}.

\textit{\textbf{Compute resources:}}
VL-T5 and VL-BART are pretrained on 4 RTX 2080 Ti GPUs, for 4 days. 
Total 30 training epochs and batch size of 320 and 600 were used for VL-T5 and VL-BART respectively.

\subsection{Type-B: Custom Layer based Deep Fusion (CLDF)}
\label{section:type2}

\begin{figure}[ht]
  \centering
  \includegraphics[width=0.9\textwidth]{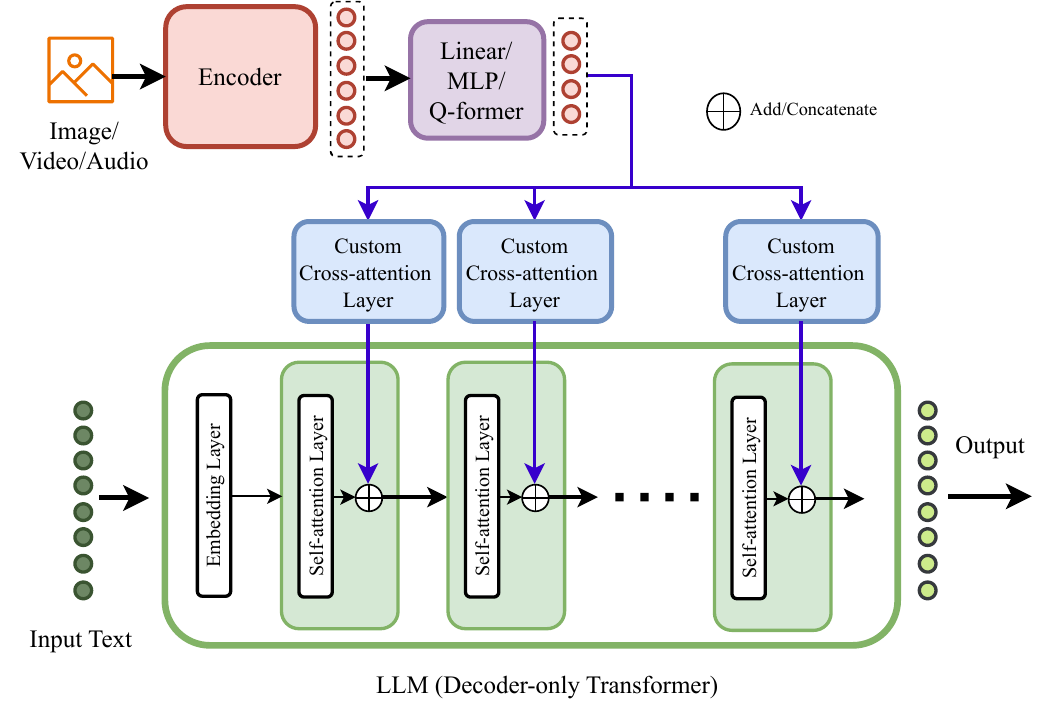}
  \caption{Type-B  multimodal model architecture. The input modalities are deeply fused into the internal layers of the LLM using custom-designed layers. Custom cross-attention layers (sub-type A.1) or other custom layers (sub-type A.2) are used for modality fusion. A Linear Layer/MLP/Q-former is used to align different modalities with the decoder layer.}
  \label{fig:type2architecture}
\end{figure}

Type-B  architecture is built using a pretrained LLM, learnable linear layer/MLP/Q-former, custom-cross-attention-layers or custom-layers and modality encoders.
The difference between Type-A and Type-B architecture types is that,
\textit{in Type-A, a standard cross-attention layer is utilized, 
while in Type-B, a custom-designed layer is or can be used}.
Similar to Type-A, in Type-B the input modalities are deeply fused into the internal layers of the model.
Example, LLaMA-Adapter-V2 model adds learnable embeddings to the output of the cross-attention layer before adding/concatenating the cross-attention output to the self-attention layer output. 
Additionally, a learnable gating factor is included to control the contribution of the cross-attention layer on the output of self-attention layer, which is not typically done in Type-A architecture.
Models belonging to Type-B include LLaMA-Adapter \cite{zhang2023llama}, LLaMA-Adapter-V2 \cite{gao2023llama}, CogVLM \cite{wang2023cogvlm}, mPLUG-Owl2 \cite{ye2023mplug}, CogAgent \cite{hong2023cogagent}, InternVL \cite{chen2023internvl}, MM-Interleaved \cite{tian2024mm}, CogCoM \cite{qi2024cogcom}, InternLM-XComposer2 \cite{dong2024internlm}, MoE-LLaVA \cite{lin2024moe}, and LION \cite{chen2023lion}.
Figure \ref{fig:type2architecture} shows general Type-B multimodal model architecture.
Two architecture sub-types exists for Type-B. 
Sub-type B.1 \ref{section:type2subtype1} adds custom-cross-attention layer to the internal layers of the LLM, while the sub-type B.2 \ref{section:type2subtype2} uses custom learnable layer other than cross-attention layer.
Sections \ref{section:type2subtype1} and \ref{section:type2subtype2} provide more details about sub-type B.1 and B.2 respectively.
Models with multimodal input and text output dominate the Type-B architecture, similar to Type-A.
If a Mixture-of-Experts layer (\cite{jacobs1991adaptive}; \cite{eigen2013learning}) is used in the architecture, an auxilary loss is added to the standard auto-regressive loss (next-text-token prediction task) for load balancing.
The comparative advantages and disadvantages of the Type-B multimodal model architecture, in relation to Types A, C, and D, are detailed in Section\ref{section:type2advantages}.

\subsubsection{Sub-type B.1: Custom Cross-Attention Layer}
\label{section:type2subtype1}

Here, the multimodal models are built through the process of using pretrained LLM and adding a \textit{custom cross-attention layer} to the internal layers of the decoder.
Models include LLaMA-Adapter \cite{zhang2023llama}, LLaMA-Adapter-v2 \cite{gao2023llama}, CogVLM \cite{wang2023cogvlm}, mPLUG-Owl2 \cite{ye2023mplug}, CogAgent \cite{hong2023cogagent}, InternVL \cite{chen2023internvl}, MM-Interleaved \cite{tian2024mm}, and CogCoM \cite{qi2024cogcom}. 

\textbf{CogVLM} learns separate query (Q), key (K), and value (V) embeddings for text and images in each decoder layer. 
These Q, K, V are initialized to same value as LLM at start of the training. 
A visual expert module processes the encoder outputs, and then inputs it to the custom cross-attention layers in the decoder (LLM).
While \textbf{mPLUG-Owl2}, introduces a custom-cross-attention layer in which a common Q is learnt for separate K and V for each modality. 
This custom-cross-attention layer is called as `Modality adaptive module' in this architecture. 
\textbf{MM-Interleaved}, introduces a custom-cross-attention layer called as MMFS (Multi-scale Multi-image Feature Synchronizer). 
This feature synchronizer is added after self-attention in each layer of the LLM.
In \textbf{LLaMA-Adapter} and \textbf{LLaMA-Adapter-V2}, a custom-cross-attention layer called as `adapter' is used. 
The Q in this cross-attention layer are learnable embeddings, and K, V are from the encoder outputs (after linear layer). 
Additional learnable embeddings are added to the output of the cross-attention before sending it to the decoder layer. 
This custom-cross-attention layer output is concatenated to the output of decoder layer.

\textit{\textbf{Training and data:}} 
\textbf{LLaMA-Adapter} \cite{zhang2023llama} adds learnable prompts to layers of LLM for instruction following and multimdoal learning. 
It uses standard next-text-token prediction objective for instruction tuning. 
The input consists of both text and image modalities, whereas the output is limited to text only.
Dataset of size 52K samples is used to instruction tune LLaMA-Adapter.
\textbf{LLaMA-Adapter-V2} \cite{gao2023llama} further extends LLaMA-Adapter to create a visual instruction model. 
The input comprises text and images, while the output consists solely of text.
This model too is trained on standard next-text-token prediction task. 
It uses 52K language instruction data \cite{peng2023instruction} \& 567K image-text captioning data from COCO caption dataset \cite{chen2015microsoft} and 80K conversation data collected by ShareGPT \cite{sharegpt2023sharegpt} for finetuning/instruction tuning. 
In \textbf{CogVLM}, an additional attention layer and FFN (Linear Layer/s) are added in parallel to the LLM's self-attention and FFN layer, for learning image features. Model takes image and text as input, and outputs text. 
Since output is text-only, this model too uses standard next-text-token prediction task for training.
Pretraining data include LAION-2B, COYO-700M \cite{kakaobrain2022coyo-700m}, LAION-115M \cite{li2023blip} and a subset
of LAION-400M is used.
Instruction tuning uses VQAv2, OKVQA, TextVQA, OCRVQA, ScienceQA \cite{lu2022learn}, LLaVA-Instruct \cite{liu2024visual}, LRV-Instruction \cite{liu2023aligning}, LLaVAR \cite{zhang2024llavar}. Flickr30K Entities \cite{plummer2015flickr30k}, Ref-COCO \cite{kazemzadeh2014referitgame}, Visual7W, VisualGenome and Grounded CoT-VQA \cite{chen2023shikra} datasets.
\textbf{mPLUG-Owl2} accepts both image and text inputs, producing text as output.
Training objective is the standard next-text-token prediction.
Pretraining datasets used are CC3M/CC12M \cite{changpinyo2021conceptual}, COCO \cite{lin2014microsoft}, Laion-en \cite{schuhmann2022laion}, COYO \cite{kakaobrain2022coyo-700m}, DataComp \cite{gadre2024datacomp}. 
Visual Instruction tuning datasets include TextCaps \cite{sidorov2020textcaps}, COCO, VQAv2, OKVQA, OCR-VQA, GQA, and A-OKVQA, Ref-COCO, VisualGenome, LLaVA-instruct-150K \cite{liu2024visual}, ShareGPT-80K \cite{sharegpt2023sharegpt}, SlimOrca \cite{SlimOrca}.
\textbf{CogAgent} is trained using standard next-text-token prediction objective. 
It processes image and text as input, and outputs text.
It is pretrained using datasets like LAION-2B, COYO, LAION-115M.
To train model for GUI grounding, CogAgent created CCS400K (Common Crawl Screenshot 400K) dataset \cite{hong2023cogagent}.
Finetuning and alignment tuning uses Mind2Web \cite{deng2024mind2web} and AITW \cite{rawles2024androidinthewild}.
\textbf{InternVL}, similar to other models in this architecture type, takes image and text as input and outputs text.
Training progresses through three stages \cite{chen2023internvl}. 
In stage 1, contrastive training is utilized to construct a 6 billion parameter vision model.
In stage 2, a standard generative training is used with InternViT and frozen QLLaMA for image captioning task.
In stage 3, supervised finetuning is done for visual QA and multimodal dialogue task.
Stage 1 and 2 uses LAION-en, LAION-multi \cite{schuhmann2022laion},
COYO, Wukong \cite{gu2022wukong}, LAION-COCO \cite{schuhmann2022laion}, LAION-en, CC12M, CC3M, SBU \cite{ordonez2011im2text} datasets.
While stage 3 uses COCO Caption, TextCaps, VQAv2, OKVQA, A-OKVQA, IconQA \cite{lu2021iconqa}, AI2D \cite{kembhavi2016diagram}, GQA, OCR-VQA, ChartQA, DocVQA, ST-VQA, EST-VQA \cite{wang2020general}, InfoVQA, LLaVAR, Toloka \cite{ustalov2023toloka}, LLaVA-150K, SVIT \cite{zhao2023svit}, VisDial \cite{das2017visual}, LRV-Instruction, LLaVA-Mix-665K \cite{liu2023improved} datasets.
\textbf{MM-Interleaved}
model is trained end-to-end with next-text-token and next-image prediction task \cite{tian2024mm}. 
The architecture is made up of visual encoder, resampler, LLM, feature-synchronizer and a diffusion model. 
Image and text are given as input, and model can output both image and text.
Model is pretrained on a mixture of image-text pairs and interleaved image-text sequences, including MMC4, LAION-2B, LAION-COCO, CC-12M and Objects365 \cite{shao2019objects365}.
Finetuning data include LLaVA-Mix-665K, COCO Caption, VQAv2, ChartQA, DocVQA, EST-VQA, InfoVQA, STVQA, TextCaps, LLaVAR, OCR-VQA, and DVQA, RefCOCO, RefCOCO+ \cite{mao2016generation}, and RefCOCOg \cite{mao2016generation}.

\textit{\textbf{Compute resources:}}  
LLaMA-Adapter uses 8 A100 GPUs. Only 1 hour of triaining required on 52K language instruction following dataset using LLaMA-7b.
LLaMA-Adapter-V2, since derived from LLaMA-Adapter, it too is trained using 8 A100 GPUs. Training time varies due to difference in training data sizes.
CogVLM, mPLUG-Owl2, CogAgent and MM-Interleaved did not explicitly include details about training resources in their work.
InternVL uses 640 A100 GPUs for stage 1 training, 160 A100 GPUs for stage 2 training and 8 to 32 A100 GPUs for stage 3 training.

\subsubsection{Sub-type B.2: Custom Learnable Layer}
\label{section:type2subtype2}

Instead of adding custom cross-attention layers to internal LLM layers, models using \textit{custom learnable layers} belong to sub-type B.2 of Type-B multimodal model architecture.
Models incude InternLM-XComposer2 \cite{dong2024internlm}, MoE-LLaVA \cite{lin2024moe} and LION \cite{chen2023lion}.
\textbf{InternLM-XComposer2} adds LoRA weights to LLM layers for learning image modality.
The LoRA weights are added in parallel to each decoder layer and only processes image tokens.
No cross-attention layer is added in the decoder layers to build this model.
\textbf{MoE-LLaVA} model is built on top of LLaVA multimodal model (LLaVA model belongs to Type-C).
In addition to the changes that exists in LLaVA model, in MoE-LLaVA each decoder layer of LLaVA is modified to create MoE-LLaVA model. 
In each decoder layer, the FFN layer is modified to create Mixture-of-Expert (MoE) layer. 
MoE layer consists of a router and multiple FFN layers in parallel. 
\textbf{LION} modifies the FFN layer of each decoder layer. 
The learnable module 'Mixture of adapters with routers' is added in parallel to the FFN layers. 
This module consists of LoRA and MoE layer for learning image modality features.

\textit{\textbf{Training and data:}}
\textbf{InternLM-XComposer2} can process image and text as input, and outputs text.
The model is trained using a standard next-text token prediction task.
Its training process involves pretraining, finetuning and instruction tuning.
Pretraining datasets include ShareGPT4V-PT \cite{chen2023sharegpt4v}, COCO, Nocaps, TextCaps, LAION-400M, SBU, CC-3M, Concept Data \cite{zhang2023internlm}, WanJuan \cite{he2023wanjuan}, Flicker, MMC-Instruction \cite{liu2023mmc}.
Finetuning data encompasses ShareGPT4V, COCO, Nocaps, VQAv2, GQA, OK-VQA, Science QA, AI2D, SQA, DVQA, ChartQA, MathQA \cite{amini2019mathqa}, Geometry3K \cite{lu2021inter}, A-OKVQA, KVQA \cite{shah2019kvqa}, LLaVA-150k, LVIS-Instruct4V \cite{wang2023see}.
Instruction tuning data include LLaVA-150k, LVIS-Instruct4V,
ShareGPT-en\&zh \cite{chiang2023vicuna}, InternLM-Chat \cite{team2023internlm}.
\textbf{MoE-LLaVA} too processes image and text as input, and outputs text.
An auxilary loss related to the mixture-of-experts load balancing is added to the standard auto-regressive loss.
Training consists of three stages: one pretraining and two finetuning stages.
Data used for pretraining is LLaVA 1.5-558k \cite{liu2023improved} dataset.
First finetuning stage uses SViT-157k \cite{zhao2023svit}, LVIS-220k, LRV-331k \cite{liu2023aligning}, and MIMIC-IT-256k datasets.
Second finetuning stage uses LLaVA 1.5-mix-665k dataset.
\textbf{LION} receives input in the form of images and text, and produces text as its output.
It uses standard next-text-token prediction objective for training.
LoRA weights and MoE layer are added to LLM layers for learning image modality.
Training data includes LLaVA-Instruct-150K , OKVQA, A-OKVQA, VQAv2, OCR-VQA, COCO, TextCaps and Visual Genome datasets.

\textit{\textbf{Compute resources:}}
All MoE-LLaVA training steps use 8 A800 (80G) GPUs.
InternLM-XComposer2 and LION does not explicitly provide resource details in their work.

\subsection{Type-C: Non-Tokenized Early Fusion (NTEF)}
\label{section:type3}

\begin{figure}[ht]
  \centering
  \includegraphics[width=0.9\textwidth]{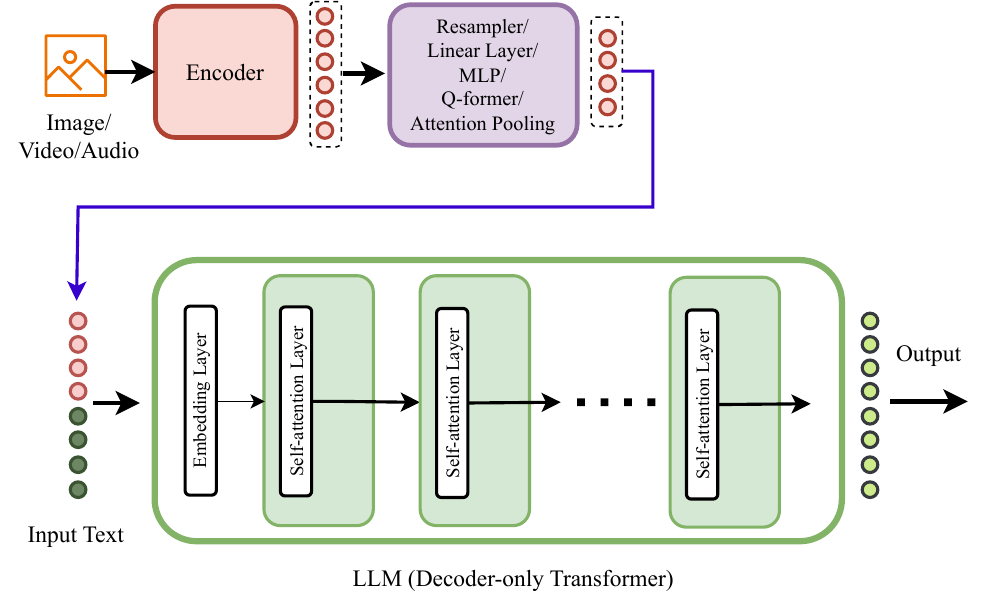}
  \caption{Type-C multimodal model architecture. The (non-tokenized) input modalities are directly fed to the model at its input, rather than to its internal layers, resulting in early fusion. Different types of modules are used to connect modality encoder outputs to the LLM (model) like a Linear-Layer/MLP (sub-type C.1), Q-former and a Linear-Layer/MLP (sub-type C.2), Perceiver resampler (sub-type C.3), Custom learnable layers (sub-type C.4).}
  \label{fig:type3architecture}
\end{figure}

Type-C stands out as the most widely adopted multimodal model architecture.
Figure \ref{fig:type3architecture} depicts a general Type-C multimodal model architecture.
The modular nature of this architecture type contributes to its simplicity in both construction and training.
Differing from Type-A and Type-B, in Type-C and Type-D architectures, the modality encoder output is solely directed and fused at the input of the model, without involvement in the internal layers of the model.
Hence, Type-C belongs to early fusion category (Figure \ref{fig:types_subtypes}).
Pretrained LLM as decoder is used without any major architectural changes to its internal layers
(some models belonging to this type may have LoRA weights added to their decoder layers).
Pretrained image encoder or other modality encoder are used. 
The encoder/s and the decoder are combined together with learnable module like single Linear-Layer, MLP, Q-former, attention-pooling layer, convolutional layer, perceiver resampler or variants of Q-former.
Q-former, first introduced in BLIP-2 \cite{li2023blip}, is a lightweight transformer architecture consisting of two components. 
One utilizes learnable query embeddings and cross-attention for images, while the other employs self-attention for text processing. 
Both components share a self-attention layer, enhancing efficiency in multimodal tasks.
The Perceiver resampler, similar to Q-former, utilizes learnable queries and cross-attention mechanisms to process image data, generating a fixed number of visual tokens. However, unlike Q-former, it does not incorporate self-attention layers specifically tailored for text processing.
Unlike Type-A and B, in Type-C, no major internal model architectural changes are added to either encoder or decoder.
This allows the multimodal model belonging to Type-C, to incorporate off-the-shelf LLMs and encoders in their architecture. 
A substantial number of models fall under Type-C. Thus, for clarity, our study categorizes them according to the type of connectors utilized to link the modality encoder and LLM.
Section \ref{section:type3subtype1}, \ref{section:type3subtype2}, \ref{section:type3subtype3} and \ref{section:type3subtype4} lists and describes the the sub-types/categories.
The comparative advantages and disadvantages of the Type-C multimodal model architecture, in relation to Types A, B, and D, are detailed in Section\ref{section:type3advantages}.

\subsubsection{Sub-type C.1: Linear Layer/MLP}
\label{section:type3subtype1}

\textbf{Models using only Linear Layer/MLP for connecting Encoder to the LLM (decoder)}:
DeepSeek-VL \cite{lu2024deepseek}, LLaVA \cite{liu2024visual}, LLaVA-Med \cite{li2024llava}, LLaVAR \cite{zhang2024llavar}, LLaVA-1.5 \cite{liu2023improved}, LLaVA-Phi \cite{zhu2024llava}, LLaVA-NeXT \cite{liu2024llavanext}, PaLM-E \cite{driess2023palm}, MiniGPT-v2 \cite{chen2023minigpt}, DetGPT \cite{pi2023detgpt}, PandaGPT \cite{su2023pandagpt}, GILL \cite{koh2024generating}, Shikra \cite{chen2023shikra}, GPT4RoI \cite{zhang2023gpt4roi}, ChatSpot \cite{zhao2023chatspot}, NExT-GPT \cite{wu2023next}, Fuyu \cite{fuyu-8b}, FROMAGe \cite{koh2023grounding}, ShareGPT4V-7B \cite{chen2023sharegpt4v}, GroundingGPT \cite{li2024groundinggptlanguage}, ModaVerse \cite{wang2024modaverse}, MLLM-Tool \cite{wang2024mllmtool}, ViGoR \cite{yan2024vigor}, CoDi \cite{tang2024any}, CoDi-2 \cite{tang2023codi}.

\subsubsection{Sub-type C.2: Q-former and Linear Layer/MLP}
\label{section:type3subtype2}

\textbf{Models using Q-former and Linear Layer/MLP for connecting Encoder to the LLM (decoder)}:
BLIP-2 \cite{li2023blip}, MiniGPT-4 \cite{zhu2023minigpt}, X-LLM \cite{chen2023x}, InstructBLIP \cite{dai2024instructblip}, EMU \cite{sun2023generative}, Video-LLaMA \cite{zhang2023video}, BLIVA \cite{hu2024bliva}, SALMONN \cite{tang2023salmonn}, X-InstructBLIP \cite{panagopoulou2023x}, BuboGPT \cite{zhao2023bubogpt}, VILA \cite{lin2023vila}, TinyGPT-V \cite{yuan2023tinygpt}, SPHINX \cite{lin2023sphinx}, SPHINX-X \cite{gao2024sphinx}.

\subsubsection{Sub-type C.3: Perceiver Resampler}
\label{section:type3subtype3}

\textbf{Models using Perceiver resampler for connecting Encoder to the LLM (decoder)}:
Idefics2 \cite{laurenccon2024matters}, InternLM-XComposer \cite{zhang2023internlm}, KOSMOS-2.5 \cite{lv2023kosmos}, Monkey \cite{li2023monkey}, V* \cite{wu2023textit}, Kosmos-G \cite{pan2024kosmosg}.

\subsubsection{Sub-type C.4: Custom Learnable layer}
\label{section:type3subtype4}

\textbf{Models using custom-module/layer for connecting Encoder to the LLM (decoder)}:
MM1 \cite{mckinzie2024mm1}, mPLUG-Owl \cite{ye2024mplugowl}, mPLUG-DocOwl \cite{ye2023mplugdocowl}, EmbodiedGPT \cite{mu2024embodiedgpt}, Video-ChatGPT \cite{maaz2023video}, Qwen-VL \cite{bai2023qwen}, AnyMAL \cite{moon2023anymal}, DocPedia \cite{feng2023docpedia}, mPLUG-PaperOwl \cite{hu2023mplug}, Osprey \cite{yuan2023osprey}, EMU2 \cite{sun2023generative}, KAM-CoT \cite{mondal2024kam}, VisLingInstruct \cite{zhu2024vislinginstruct}, MobileVLM \cite{chu2023mobilevlm}, MobileVLM-V2 \cite{chu2024mobilevlm}.

\subsubsection{Training Methods and Data}
Many recent multimodal survey works like \cite{zhang2024mmllms}, \cite{yin2024survey}, \cite{guo2023survey}, \cite{caffagni2024r} and \cite{wu2023multimodal} have explored Type-C multimodal model architecture in great detail.
This section will highlight some important details provided in these work here, that are related to model training strategies, data and resources.

\textit{\textbf{Training and data:}}
Three stages of training used for Type-C multimodal model architecture are pre-training, instruction-tuning, and alignment tuning.
General training procedure (each model may have some variations): 
Step 1 (Pretraining): Freeze LLM and Encoder. 
Only train the projection layer/s for Vision-Language alignment. 
Step 2 (Instruction and alignment tuning): Train projection layer and LLM for multimodal tasks. 
The encoder is trained optionally if required. 
But in general, only projection layer and LLM are trained.
Pretraining mainly aims to align different modalities and learn multimodal world knowledge typically through caption data (text) for images, audio and videos \cite{yin2024survey}.
Since the output modality is text, standard next-text-token prediction objective is used during pretraining.
Table \ref{table:type3datapretraining} lists pretraining data commonly used for Type-C architecture.
It contains captioning data for images, audio and video modalities.
LLMs and Multimodal-LLMs are being widely deployed in real-world applications, specially in chat-based applications (chat-bots).
These applications require the model to understand user query (instruction) and with a greater attention on details provided by the user.
In order to better align model for query (instruction) understanding, models are trained on instruction following datasets.
The training process is called as instruction tuning.
Instruction following datasets contain wide variety of tasks, hence making model more versatile across different tasks.
This training process not only increases the instruction following capability, but also leads to improved few-shot and zero-shot performances.
The survey works \cite{zhang2024mmllms} and \cite{yin2024survey} lists different ways used to instruction tune a multimodal model of Type-C.
Table \ref{table:type3instructiontuningdata} shows the datasets used for instruction tuning of Type-C models.
To further align model for human interactions (chat applications), the model is trained using RLHF (\cite{christiano2017deep}; \cite{ziegler2019fine}; \cite{stiennon2020learning}; \cite{bai2022training}; \cite{ouyang2022training}) or DPO \cite{rafailov2024direct} training strategies using human preference data.
Table \ref{table:type3alignmentdata} enumerates datasets used for model alignment.

\textit{\textbf{Compute resources:}}
Type-C multimodal model architecture is data and compute efficient. Hence, low training resources are required compared to all other types.

\textbf{LLaVA} use 8 A100s. Pretrained for 4 hours and finetuned within 10 hours.
\textbf{LLaVA-Med} takes 7 and 8 hours for stage 1 and 2 training on 8 (40G) A100 GPUs.
\textbf{LLaVAR} all experiments are run on NVIDIA A100 (80GB) GPUs.
\textbf{LLaVA-1.5} full training completed within 1 day on a single 8-A100 node.
\textbf{LLaVA-Phi} uses 8 A100 GPUs. Pretrained for 1.5 hours. 8 hours for visual instruction tuning.
\textbf{LLaVA-NeXT} full trains within approximately 1 day with 32 A100s GPUs.
\textbf{MiniGPT-v2} stage 1 training requires 8 A100 GPU for around 90 hours. 
Second stage is trained on 4 A100 GPU for roughly 20 hours.
Last stage, training is executed on 4 A100 GPUs for around 7 hours.
\textbf{PandaGPT} uses 8 A100 (40G) GPUs for around 7 hours for training.
GILL training utilizes 2 A6000 GPUs for 2 days.
\textbf{Shikra} model's all training runs on 8 NVIDIA A100 GPUs. It takes around 100 hours for stage one training and 20 hours for stage two.
All \textbf{FROMAGe} training steps are completed within 1 day (24 hours) using a single A6000 GPU.
\textbf{DeepSeek-VL} 7B model utilized a cluster of 64 nodes, each comprising 8 Nvidia A100 GPUs for 5 days, while DeepSeek-VL-1B consumed 7 days with 16 nodes for all training steps \cite{lu2024deepseek}.
\textbf{ModaVerse} model's full training completed in 20 hours on 4 A100 GPUs.

\textbf{BLIP-2} uses  16 A100 (40G) GPUs.
In \textbf{MiniGPT-4},  stage 1 vision-language alignment training uses 4 A100 GPUs for 10 hours. And stage 2 finetuning takes 7 minutes on single A100 GPU.
All \textbf{X-LLM} experiments use up to 8 A100 (40G) GPUs. 
\textbf{InstructBLIP} models are trained utilizing 16 Nvidia A100 (40G) GPUs for 1.5 days.
\textbf{EMU} pretraining uses 128 NVIDIA A100 (80G) GPUs. While, instruction tuning stage utilizes 16 A100 (80G) GPUs.
\textbf{X-InstructBLIP} requires 8 A100 (40GB) GPUs.
\textbf{TinyGPT-V} can be trained on a 24GB memory GPU. Phi-2 and CLIP are used as LLM and vision encoder.
\textbf{SPHINX} pre-trains for 125 hours on 32 A100 GPUs with a 7B language model and about takes twice amount of time for 13B language model.  
Fine-tuning takes about 38 hours with 16 A100 GPUs with a 13B language model.

\textbf{InternLM-XComposer} uses 128 Nvidia A100 GPUs.
\textbf{Monkey} model's whole training process takes 40 A800 days for one epoch.
For \textbf{KOSMOS-G}, the training process took around 4 days with 256 NVIDIA V100 GPUs.
No resource details are provided in KOSMOS-2.5, V* and idefics2 works.

\textbf{Video-ChatGPT} 7B model training completed in 3 hours on 8 A100 (40GB) GPUs.
\textbf{AnyMAL} was able to train a 70B model on a single A100 (80GB) VRAM GPU through quantization. 
Other models used a varying number of Nvidia A100 GPUs.
\textbf{DocPedia} used 8 A100 GPUs for training.
\textbf{mPLUG-PaperOwl} model is trained equivalent to costing 64 A100 days.
\textbf{Osprey}'s training is conducted on 4 NVIDIA A100 GPUs with 80GB memory.
In \textbf{MobileVLM}, first the language model pretraining is completed using 20 nodes equipped with 8 NVIDIA Tesla A100 GPUs each. Later, vision-language training is done in 5 hours with 8 NVIDIA Tesla A100 GPUs for MobileVLM 1.7B, and 8 hours for MobileVLM 3B.
\textbf{MobileVLM-v2}, trained on top of MobileVLM, is pretrained on 8 NVIDIA A100 GPUs for about 5 hour. 
Finetuning is peformed with 8 NVIDIA A100 GPUs for around 9 hours.
Resource information are not clearly provided in MM1, mPLUG-Owl, mPLUG-DocOwl, EmbodiedGPT, Qwen-VL and EMU2.

\begin{table}[ht]
  \caption{Pretraining data for Type-C. \cite{yin2024survey}, \cite{wu2023multimodal}, \cite{zhang2024mm} }
  \label{table:type3datapretraining}
  \centering
  \begin{tabular}{lcc}
    \toprule
    Dataset     &  Modality & Samples    \\
    \midrule
    ALLaVA & $I+T \rightarrow T$ & 709K       \\
    LVIS-Instruct4V & $I+T \rightarrow T$ & 111K  \\
    ShareGPT4V-PT & $I+T \rightarrow T$ & 1.2M  \\
    COYO-700M & $I+T \rightarrow T$ & 747M       \\
    LAION-COCO & $I+T \rightarrow T$ & 600M  \\
    LAION-2B & $I+T \rightarrow T$ & 2.3B  \\
    LAION-5B  & $I+T \rightarrow T$   & 5.9B  \\
    CC-12M  & $I+T \rightarrow T$   & 12.4M  \\
    CC-3M  & $I+T \rightarrow T$  & 3.3M  \\
    SBU Captions  & $I+T \rightarrow T$   & 1M  \\
    \midrule
    VTP & $V+T \rightarrow T$ & 27M  \\
    WebVid2M &  $V+T \rightarrow T$ & 2.5M  \\
    YouCook2  & $V+T \rightarrow T$ & 2.2K \\
    MSR-VTT & $V+T \rightarrow T$ & 200K  \\
    \midrule
    AISHELL-1 & $A+T \rightarrow T$ & 128K  \\
    AISHELL-2 & $A+T \rightarrow T$ & 1M  \\
    WavCaps & $A+T \rightarrow T$ & 24K  \\
    Common Voice & $A+T \rightarrow T$ & 9.2K  \\
    LibriSpeech & $A+T \rightarrow T$ & 1K  \\
    
    \midrule
    MM5Product & $I+A+V+T$ &  6M  \\
    MSR-VTT & $I+A+V+T$ &   10K  \\
    
    \bottomrule
  \end{tabular}
\end{table}

\begin{table}[ht]
  \caption{Instruction tuning data for Type-C. \cite{zhang2024mmllms}, \cite{yin2024survey}}
  \label{table:type3instructiontuningdata}
  \centering
  \begin{tabular}{lcc}
    \toprule
    Dataset     &  Modality & Samples      \\
    \midrule
    LLaVA-Instruct & $I+T \rightarrow T$ & 158K \\
    LVIS-Instruct & $I+T \rightarrow T$ & 220K \\
    ALLaVA & $I+T \rightarrow T$ & 1.4M  \\
    \midrule
    MIMIC-IT   &  $ I/V +T \rightarrow T $  &   2.8M  \\ 
    Video-ChatGPT & $V+T \rightarrow T$ & 100K  \\
    VideoChat & $V+T \rightarrow T$ & 11K \\
    \midrule
    Clotho-Detail & $A+T \rightarrow T$ & 3.9K  \\
    BuboGPT’s IT & $(I+A)/A+T \rightarrow T$ & 9K \\
    \midrule
    T2M & $T \rightarrow I/V/A+T $ & 14.7K  \\
    \midrule
    MosIT & $I+V+A+T \rightarrow I+V+A+T $ & 5K  \\
    
    \bottomrule
  \end{tabular}
\end{table}

\begin{table}[h!]
  \caption{Alignment tuning data for Type-C. \cite{zhang2024mmllms}}
  \label{table:type3alignmentdata}
  \centering
  \begin{tabular}{lcc}
    \toprule
    Dataset     &  Modality & Samples     \\
    \midrule
    VLGuard’s IT  & $I+T \rightarrow T$ & 3K    \\
    RTVLM & $I+T \rightarrow T$ & 5K  \\
    MMViG & $I+T \rightarrow T$ & 16K \\
    LLaVA-RLHF  & $I+T \rightarrow T$ & 10K  \\
    RLHF-V’s IT   & $I+T \rightarrow T$ & 1.4K \\

    \bottomrule
  \end{tabular}
\end{table}

\subsection{Type-D: Tokenized Early Fusion (TEF)}
\label{section:type4}

\begin{figure}[ht]
  \centering
  \includegraphics[width=0.9\textwidth]{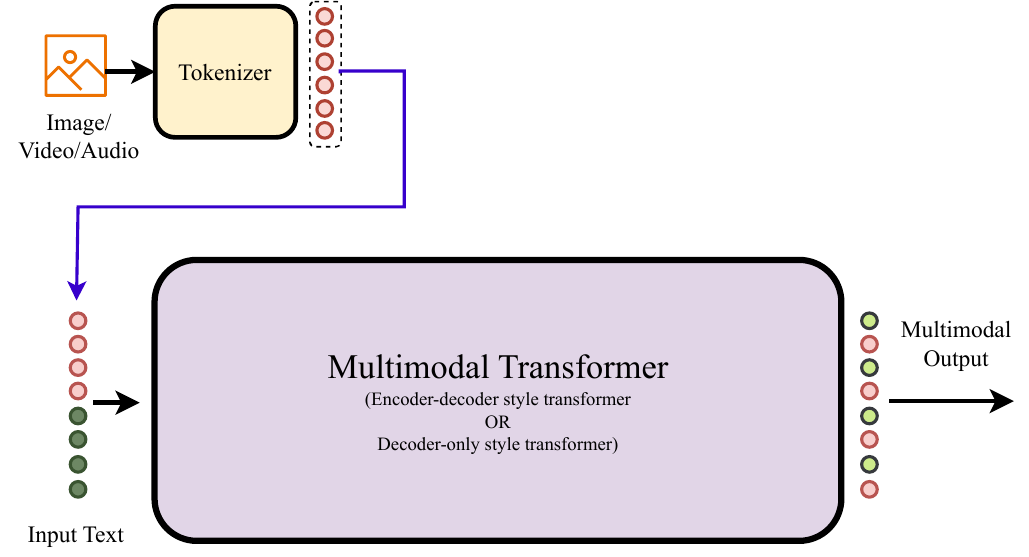}
  \caption{Type-D multimodal model architecture. The tokenized input modalities are directly fed into the model at its input. Either a decoder-only transformer (sub-type D.1) or an encoder-decoder style transformer (sub-type D.2) is used a the multimodal transformer in this architecture.}
  \label{fig:type4architecture}
\end{figure}

In Type-D, multimodal inputs are tokenized using a common tokenizer or modality specific tokenizers. 
The tokenized inputs are then given to a pretrained LLM \ref{subsubsection:type4_1} or an encoder-decoder transformer model \ref{subsubsection:type4_2}, which generates multimodal outputs.
Figure \ref{fig:type4architecture} represents a general Type-D multimodal model architecture.
Either pretrained modality specific tokenizers are used, or a tokenizer training stage is included in the training process.
The fundamental advantage of tokenizing the inputs is that, the model now can be trained auto-regressively to generate image, audio and different modality tokens along with text tokens.
Models belonging to Type-D include, LaVIT \cite{jin2024unified}, TEAL \cite{yang2024teal}, CM3Leon \cite{yu2023scaling}, VL-GPT \cite{zhu2023vl}, Unicode \cite{zheng2024unicode}, SEED \cite{ge2023making}, 4M \cite{mizrahi20244m}, Unified-IO \cite{lu2022unified}, Unified-IO-2 \cite{lu2023unified}.
The comparative advantages and disadvantages of the Type-D multimodal model architecture, in relation to Types A, B, and C, are detailed in Section\ref{section:type4advantages}.

\subsubsection{Subtype D.1: Models using LLM}
\label{subsubsection:type4_1}

Models that primarily use LLM are LaVIT, TEAL, CM3Leon, SEED, Unicode, VL-GPT.
\textbf{LaVIT} aims at unified generative training for image and text modalities.
It is achieved using a visual tokenizer.
After image encoding using an image encoder, a visual tokenizer is used in LaVIT model architecture to tokenize the visual inputs.
\textbf{TEAL}, tokenizes all modalities. 
It has a tokenizer and a detokenizer module in its architecture. 
A projection layer is used for connecting non-textual modalities to the LLM. 
\textbf{CM3Leon} uses image tokenizer to tokenize images and then directly provide it to the LLM. 
It uses OPT model as a LLM.
In \textbf{VL-GPT}, a tokenizer is first trained to convert images to image tokens.
Later, the trained tokenizer is used to feed the images to the LLM.

\textit{\textbf{Training and data:}}
\textbf{LaVIT} model is trained in two stages. 
In stage I, a image tokenizer is trained. 
In stage II, the main LaVIT model is trained.
LaVIT utilizes general auto-regressive objective function where likelihood of each multi-modal sequence is directly maximized \cite{jin2024unified}.
Since both image and text are already represented as discrete tokens, the cross-entropy loss is used to supervise the token prediction at each location for both modalities (image and text) with a shared prediction head \cite{jin2024unified}.
\textbf{TEAL} employs pretrained tokenizers. 
During pretraining of TEAL, only the projection layers are trained.
The LLM and the encoders are frozen for textual and non-textual embedding alignment during pretraining. 
It uses image-text and audio-text pairs for pretraining. Finetuning is done on downstream tasks. 
Datasets like COCO-Caption, Science-QA, and CoVoST 2 \cite{wang2020covost} are used. 
It uses general auto-regressive objective for training.
\textbf{CM3Leon} too uses standard next-token prediction loss (auto-regressive). 
For pretraining, LAION and licensed images from shuttershock are used. 
Finetuning uses COCO Captioning (2015), Flickr30k, Image Paragraph \cite{krause2017hierarchical}, Localized Narratives \cite{pont2020connecting}, VQA2, VizWiz, OKVQA, ScienceQA, and InstructPix2Pix \cite{brooks2023instructpix2pix} datasets.
\textbf{VL-GPT}, also tokenizes the inputs, hence it too is able to use standard auto-regressive objective function for training. 
Pretraining involves utilizing both image-text pairs and interleaved image-text sequences.
CC3M, LAION-Aestheics, LAION-COCO,  Multimodal-C4 (MMC4) and OBELICS datasets are used for pretraining. 
Later, the model is instruction tuned using LLAVA data \cite{liu2024visual}, SVIT, COCO Caption, InstructPix2Pix and Magicbrush \cite{zhang2024magicbrush} datasets.

\textit{\textbf{Compute resources:}}
For LaVIT, 64 A100 GPUs and 12 hour training required to train tokenizer. 256 A100 GPUs and 36 hour training required for pretraining full LaVIT model.
TEAL \cite{yang2024teal} is fully trained with 8 A100 GPUs.  
CM3Leon \cite{yu2023scaling} was pretrained on 2 trillion tokens with 256 or 512 A100 GPUs and finetuned on 30 billion tokens with 64 or 128 A100 GPUs.
In VL-GPT \cite{zhu2023vl}, tokenizer training used 8 NVIDIA A100 (40G) GPUs for 10,000 iterations with batch size of 1024.
Pretraining utilized 32 GPUs for 20,000 iterations with batch size of 4096 and instruction tuning was performed with 4 GPUs for 10,000 iterations using batch size of 512.

\subsubsection{Subtype D.2: Models using Encoder-Decoder style Transformer}
\label{subsubsection:type4_2}

Models using encoder-decoder style transformer instead of LLM are Unified-IO, Unified-IO 2 and 4M.
In \textbf{Unified-IO} \cite{lu2022unified} and \textbf{Unified-IO-2} \cite{lu2023unified}, an encoder-decoder style transformer is used.
The input modalities are tokenized using VQ-GAN style tokenizers.
To tokenize images and dense structures, VQ-GAN is employed. 
For audio, ViT-VQGAN is utilized.
While these models may bear resemblance to Type-C, they diverge notably. The critical distinction lies in Type-D models' utilization of discrete input and output modality-specific tokens.
The \textbf{4M} model is a multimodal model capable of processing text, RGB images, depth, normals, semantic segmentation maps, and CLIP feature maps.
In contrast to Unified-IO which uses VQ-GAN style tokenizers to tokenize all modalities, 4M utilizes modality specific tokenizers.
Here, text is tokenized using WordPiece, and VQ-VAE is used to tokenize image and image-like modalities.

\textit{\textbf{Training and data:}}
\textbf{Unified-IO-2} is trained from scratch on a large multimodal pre-training data corpus from diverse sources with a multimodal Mixture of Denoisers (MoD) objective \cite{lu2023unified}.
UNIFIED-IO-2 contains 7 billion parameters and is pre-trained from scratch on an extensive variety of multimodal data – 1 billion image-text pairs, 1 trillion text tokens, 180 million video clips, 130 million interleaved image \& text, 3 million 3D assets, and 1 million agent trajectories \cite{lu2023unified}. 
The model is instruction-tuned with a massive multimodal data by combining more than 120 datasets covering 220 tasks across vision, language, audio, and action \cite{lu2023unified}.
\textbf{4M} uses MultiMAE \cite{bachmann2022multimae}  pretraining strategy, where it takes small set of tokens from all modalities at its input, and performs cross-modal prediction coding. 
VQ-VAE is used for tokenizing image related modalities and WordPiece for text tokenization.
Conceptual Captions 12M (CC12M) is used for pretraining. 
Finetuning datasets include ImageNet-21K, ImageNet-1K, COCO detection, ADE20K, and NYUv2.

\textit{\textbf{Compute resources:}}
4M is pretrained on 500 billion tokens.
Both pretraining and finetuning utilizes 64 or 128 A100 GPUs.
For Unified-IO and Unified-IO-2, Google's TPU are used. No other resource related details are provided in their work.

\section{Advantages and Disadvantages}
\label{section:advantages_all}

\subsection{Type-A}
\label{section:type1advantages}

Type-A (\ref{section:type1}) multimodal model architecture enables fine-grained control of how modality information flows in the model.
It is end-to-end trainable and omits design of custom layers by using standard learnable layers of transformers.
Example, standard cross-attention layer is used to fuse modalities in Type-A, while in Type-B (\ref{section:type2}), a specially designed layer is used to fuse modalities.
Type-A requires large number of training data samples and computational resources as compared to Type-B and Type-C multimodal model architectures.
Challenging to build model of this architecture type, due to the prerequisite understanding of the internal layers of the LLM.
Architecture is difficult to scale as compared to Type-C (\ref{section:type3}),
especially if pretraining step is involved, because of the large number of training parameters and computational requirements.
Adding more modalities is challenging, because in Type-A, after adding image modality cross-attention layer to the LLM layer, adding other modalities to each LLM layer is difficult and has not been explored in current literature to best of our knowledge.
Type-B addresses this challenge with a gating mechanism, which allows direct addition of input modalities to the output LLM layers.
The gating mechanism involves a single learnable parameter, a multiplication and an addition operation. 
The input modality is multiplied with the learnable parameter determining the contribution of the modality, later, its output is directly added to the LLM layer output.
Number of trainable parameters can be large, due to addition of learnable cross-attention layer in each LLM layer.
Unlike Type-D \ref{section:type4}, which accommodate an autoregressive training objective across diverse modalities, Type-A encounter increased complexity when applying a standard autoregressive training objective to modalities beyond text.

\subsection{Type-B}
\label{section:type2advantages}
Similar to Type-A (\ref{section:type1}), Type-B (\ref{section:type2}) also benefits from fine-grained control of how modality information flows in the model.
It is end-to-end trainable. 
In contrast to Type-A, custom-designed layers are used in Type-B.
The custom design adds to more fine-grained control of modality fusion.
Compared to Type-A, the efficient custom design of the layers and the architecture of the model mitigate the need for extensive training data samples and computational resources.
Building these model require the knowledge of the internal layers of the LLM.
In contrast to Type-A, Type-B architecture is more scalable, due to customizable nature and computational efficiency of the custom learnable connector layers.
However, scaling may still present challenges in comparison to Type-C (\ref{section:type3}) architecture.
Adding more modalities is simplified compared to Type-A.
In Type-A, the addition of further modalities to each LLM layer becomes significantly more challenging after the inclusion of an image modality cross-attention layer, an area that remains underexplored in the literature.
The Type-B, uniquely provides an alternative by introducing a gating mechanism which can be utilized to add other modalities.
The gating mechanism enables direct addition of input modalities to the output LLM layers.
Number of trainable parameters can be controlled by design of efficient custom connector layers, hence Type-B can be efficient in terms of number of trainable parameters.
In contrast to Type-D \ref{section:type4}, where an autoregressive training objective is readily applicable across various modalities, the implementation of a standard autoregressive training objective in Type-B for non-textual modalities presents greater complexity.

\subsection{Type-C}
\label{section:type3advantages}
Type-C \ref{section:type3} architecture is modular.
Parts of the model architecture can be swapped and the resulting new model can trained efficiently for multimodal tasks.
Unlike Type-A \ref{section:type1} and Type-B \ref{section:type2}, Type-C does not have fine-grained control of how modality information flows in the model.
Different modality inputs are fused only at the input of decoder (LLM).
It is end-to-end trainable. 
Type-C requires less training data and computational resources as compared to Type-A, B and D \ref{section:type4} multimodal model architectures.
It is easier to build compared to all other type of multimodal architectures, owing to its modular architecture.
Unlike Type-A and Type-B architectures, Type-C models do not necessitate detailed knowledge of the internal layers of the LLM. Instead, only the interface details of the LLM or encoder being newly integrated are required.
Type-C architecture is scalable due to its modular design, reduced training data requirements, and computational efficiency.
Adding more modalities is easier in Type-C, compared to Type-A, B and D.
A simple learnable Linear/MLP/Q-former/custom layer can be added between the modality encoder and the LLM and trained efficiently for augmenting different modalities.
Number of trainable parameters is least in Type-C compared to Type-A, B and D. 
Hence, it is compute resource efficient from training perspective.
Unlike Type-D, where an auto-regressive objective can be used to train different modalities, here in Type-C, it is challenging to utilize standard auto-regressive objective for modalities other than text (language). 
Type-C provides an alternate way to Type-D, for any-to-any multimodal model development due elimination of input modality tokenizers, its modular architecture, training efficiency and end-to-end trainable nature.

\subsection{Type-D}
\label{section:type4advantages}
Type-D \ref{section:type4} has a simplified model architecture due to tokenization of input and output modalities, when compared to Type-A \ref{section:type1}, B \ref{section:type2} and C \ref{section:type3}.
It tokenizes all modalities.
This characteristic can be perceived as both advantageous and disadvantageous. Tokenization offers the advantage of enabling all modality training through a standard auto-regressive objective function. However, the challenge lies in training a universal tokenizer or modality-specific tokenizers.
In other architecture types, output embeddings from modality encoders are directly provided to the LLM without (discrete) tokenization.
Unlike Type-A and Type-B, Type-D does not have fine-grained control of how modality information flows in the model.
Different modality inputs are fused only at the input of the main transformer model. 
This main transformer can be a decoder (LLM) or a encoder-decoder style transformer.
It is end-to-end trainable. 
It requires large training data and computational resources as compared to Type-A, B and C multimodal model architectures.
Type-D model architecture are comparatively easier to construct than Type-A and Type-B models but is more complex to build compared to Type-C architecture type.
Type-D architecture is scalable, due to tokenization of modalities.
Incorporating additional modalities poses challenges, as it necessitates training a new tokenizer for each new modality or adapting an existing multimodal tokenizer, which can be a non-trivial undertaking.
All the modalities are learnt by the main transformer, and to add another modality or task, the model has be trained, and efficient training strategies for this type of model architecture have not been extensively investigated in the literature, to the best of our knowledge.
Adding additional modalities to the model architecture is simplest in the Type-C, when compared to Type-A, B and D.
Type-D has large number of trainable parameters compared to Type-A, B and C.
Since the LLM or transformer has to learn new modality tokens, the model has to be trained, and training in Type-D is computationally intensive.
In Type-D, an auto-regressive objective can be used to train different (all) modalities.
Like Type-C, Type-D  provides a way for any-to-any multimodal model development, enabled by its tokenization of modalities and simplified training objective.

\section{Next Generation Multimodal Architectures}

\begin{figure}[ht]
  \centering
  \includegraphics[width=1.0\textwidth]{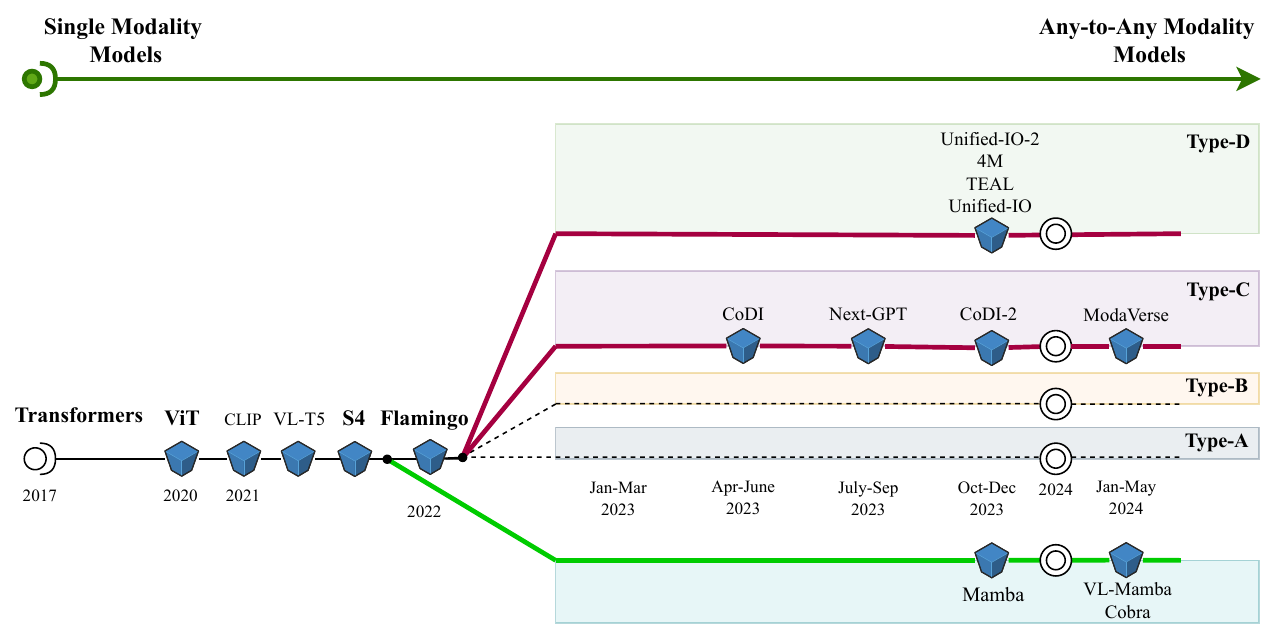}
  \caption{Any-to-any Multimodal Model development timeline. The evolution from single modality models (left) to any-to-any modality models (right) is depicted. Any-to-any multimodal models belonging to Type-C and Type-D are noted in the figure. An alternate development timeline (green line at the bottom) for non-transformer based models like SSM (State-space models) is shown. Mamba is a language model. VL-mamba and Cobra are vision-language models.}
  \label{figure:anytoany_multimodalmodelsonly}
\end{figure}

This section explores multimodal models with multimodal-input and multimodal-output.
Plethora of models exist for any-input-modality to output-text-modality.
In contrast, there are significantly fewer multimodal models capable of generating output modalities other than text.
Multimodal output generation is one of the primary challenge in the multimodal domain.
Type-C and Type-D multimodal architectures are at the forefront of development for any-to-any multimodal models.
The representative models are highlighted in Figure \ref{figure:anytoany_multimodalmodelsonly}.
These dominant multimodal model architectures address some, though not all, challenging aspects of multimodal generation.
Type-D simplifies the training process by utilizing input tokenization, enabling the use of a standard auto-regressive objective function for model training.
However, it still faces limitations in addressing the challenge of accommodating large data sizes and computational demands necessary for building multimodal generative models.
Type-C tackles the challenges associated with data and resources by leveraging pretrained components and integrating them with efficient connectors/adapters.
Yet, the training process remains challenging because of the diverse objective functions associated with different components in the model architecture.

At present, there are three primary approaches for constructing any-to-any multimodal models: the first involves utilizing the end-to-end trainable Type-D model architecture, the second entails leveraging the end-to-end trainable Type-C architecture, and the third method employs a combination of Type-C with agents, which is non-end-to-end trainable.
\textbf{Type-D architecture}: Models generate multimodal outputs using tokenizers.
Unified-IO, Unified-IO 2 and 4M are the models from Type-D which enable any-to-any multimodal model development.
\textbf{Type-C architecture}: Models generate multimodal outputs without using tokenizers.
NExt-GPT, CoDI and CoDI-2 models belong to Type-C assisting in any-to-any multimodal model development.
\textbf{Type-C + agents:} 
In this method, a Type-C multimodal model is trained to generate specific text outputs with a general format to aid the frozen pretrained modality decoder models (like text-to-image models, text-to-video models) for multimodal generation. 
ModaVerse follows this process for creating an any-to-any multimodal model. 
The absence of an end-to-end training process results in performance that is not superior to the other two methods.
Table \ref{table:compare_uIO2_codi} and \ref{table:compare_codi_next_moda} compares performance of any-to-any multimodal models.

\begin{table}[h!]
  \caption{Comparing two next generation any-to-any models. \cite{lu2023unified}}
  \label{table:compare_uIO2_codi}
  \centering
  \begin{tabular}{cccccccc}
    \toprule
    \multicolumn{1}{c}{Model} & \multicolumn{2}{c}{Image generation} & \multicolumn{4}{c}{Audio generation \& captioning} & \multicolumn{1}{c}{Video captioning} \\
    \cmidrule(lr){2-3} \cmidrule(lr){4-7} \cmidrule(lr){7-8}
    & FID↓ & TIFA↑ & FAD↓ & IS↑ & KL↓ & CIDEr↑ & CIDEr↑  \\
    \toprule
    UnifiedIO-2 & 13.39 & 81.3 & 2.64 & 5.89 & 1.80 & 48.9 & 48.8    \\
    CoDI & 11.26 & 71.6 & 1.80 & 8.77 & 1.40 & 78.9 & 74.4 \\
    \bottomrule
  \end{tabular}
\end{table}

\begin{table}[h!]
  \caption{Comparing next generation any-to-any models. Generation and captioning metrics for each modality are captured in the table. \cite{wang2024modaverse}}
  \label{table:compare_codi_next_moda}
  \centering
  \begin{tabular}{ccccccc}
    \toprule
    \multicolumn{1}{c}{Model} & \multicolumn{2}{c}{Image } & \multicolumn{2}{c}{Audio} & \multicolumn{2}{c}{Video} \\
    \cmidrule(lr){2-3} \cmidrule(lr){4-5} \cmidrule(lr){6-7}
    & FID↓ & CIDEr↑ & IS↑  & CIDEr↑  & CLIPSIM↑ & METEOR↑  \\
    \toprule    
    CoDI & 11.26 & 149.9 & 8.77 & 0.789 &  0.2890 & 32.5 \\
    NExT-GPT  & 11.28 & 156.7 & 8.35 & 0.802 & 0.3085 & 38.5 \\
    ModaVerse  & 11.24 & 151.4 & 8.22 & 0.792 &  0.3014 & 35.2 \\
    
    \bottomrule
  \end{tabular}
\end{table}

State space models (SSMs) \cite{gu2021combining}, \cite{smith2023simplified}, \cite{gu2023mamba}, are emerging as a viable alternative to transformers based model architectures.
They tackle the inherent quadratic complexity of attention mechanisms in Transformers.
Examples such as VL-Mamba \cite{qiao2024vl} and Cobra \cite{zhao2024cobra} serve as compelling illustrations of how SSMs can be extended to incorporate multimodal learning capabilities.
Their architecture closely resembles that of Type-C multimodal model architectures; therefore, it has been incorporated into the Type-C section in the Figure \ref{fig:developement_timeline}.
Moreover, works such as MambaTalk \cite{xu2024mambatalk} and SpikeMba \cite{li2024spikemba} are enhancing SSMs by incorporating modalities such as audio and video, respectively.
While any-to-any multimodal models based on SSMs have not yet been developed, the potential exists to construct such models.
Therefore, in the future, SSMs may emerge as a robust alternative to Transformer-based Type-C and Type-D multimodal model architectures for any-to-any multimodal tasks.

\section{Conclusion}

This work uniquely identifies and categorizes existing multimodal model architectures in four types.
Each architecture type is discussed in detail, including visualization of the general model architecture along with insights into their respective characteristics.
Through a thorough examination of existing architectural patterns used in the development of any-to-any multimodal models, this research effort sheds light on the two prevalent approaches (Type-C and Type-D) that are currently driving advancements in this field.
By comparing \& contrasting architecture types to each other by describing their advantage \& disadvantages, this work aids in model choices.
This study map a broad spectrum of existing multimodal models to the four identified types.
Though the model list is comprehensive, it is not exhaustive.
By establishing a taxonomy of multimodal architectures, we can effectively track and capture the evolving trends and advancements within the multimodal domain.

\bibliography{Styles/neurips_2024}

\end{document}